\documentclass[sigconf]{aamas}  % do not change this line!

\usepackage{booktabs}
\usepackage{graphicx}
\usepackage[export]{adjustbox}
\usepackage{tikz}
\usepackage{xcolor}
\usepackage{amsmath,amssymb,mathtools,bm,etoolbox}
\usepackage[normalem]{ulem}
\usepackage{lipsum}
\usepackage{multirow}
\usepackage{multicol}
\usepackage{makecell}
\usepackage{stackengine}
\usepackage{array}
\usepackage[labelformat=simple]{subcaption}

\usepackage{algorithm}
\usepackage[noend]{algpseudocode}
\usepackage[colorinlistoftodos]{todonotes}
\usepackage[font=small]{caption}
\usepackage{array}
\usepackage{ragged2e}
\makeatletter
\def\@copyrightspace{\relax}
\def\BState{\State\hskip-\ALG@thistlm}
\renewcommand{\ALG@beginalgorithmic}{\small}
\renewcommand{\paragraph}{%
  \@startsection{paragraph}{4}%
  {\z@}{.1ex \@plus 1ex \@minus .2ex}{-1em}%
  {\normalfont\normalsize\bfseries}%
}
\makeatother
\useunder{\uline}{\ul}{}
\usetikzlibrary{positioning}
\usetikzlibrary{arrows,calc,fit}
\DeclareMathOperator*{\argmax}{arg\,max}
\newcolumntype{C}[1]{>{\centering\arraybackslash}p{#1}}
\newcolumntype{M}[1]{>{\centering\arraybackslash}m{#1}}
\newcolumntype{R}[1]{>{\RaggedLeft\arraybackslash}p{#1}}

\setcopyright{ifaamas}  % do not change this line!
\acmDOI{}  % do not change this line!
\acmISBN{}  % do not change this line!
\acmConference[AAMAS'18]{Proc.\@ of the 17th International Conference on Autonomous Agents and Multiagent Systems (AAMAS 2018)}{July 10--15, 2018}{Stockholm, Sweden}{M.~Dastani, G.~Sukthankar, E.~Andr\'{e}, S.~Koenig (eds.)}  % do not change this line!
\acmYear{2018}  % do not change this line!
\copyrightyear{2018}  % do not change this line!
\acmPrice{}  % do not change this line!

\begin{document}

\title{A Deep Policy Inference Q-Network for Multi-Agent Systems}  % put your title here!
%\titlenote{Produces the permission block, and copyright information}

% AAMAS: as appropriate, uncomment one subtitle line; see camera ready instructions
%\subtitle{Extended Abstract}
%\subtitle{Industrial Applications Track}
%\subtitle{Socially Interactive Agents Track}
%\subtitle{Blue Sky Ideas Track}
%\subtitle{Robotics Track}
%\subtitle{JAAMAS Track}
%\subtitle{Doctoral Mentoring Program}

%\subtitlenote{The full version of the author's guide is available as \texttt{acmart.pdf} document}

% replace this with your author block!
\author{Zhang-Wei Hong$^*$, Shih-Yang Su$^*$, Tzu-Yun Shann$^*$, Yi-Hsiang Chang$^*$, Chun-Yi Lee$^*$}
\thanks{* indicates equal contribution.}
\affiliation{National Tsing Hua University\\
  \{williamd4112, at7788546, arielshann\}@gapp.nthu.edu.tw,\\
  s106062520@m106.nthu.edu.tw,
  cylee@cs.nthu.edu.tw}

\begin{abstract}  % put your abstract here!
We present DPIQN, a deep policy inference Q-network that targets multi-agent systems composed of controllable agents, collaborators, and opponents that interact with each other. We focus on one challenging issue in such systems---modeling agents with varying strategies---and propose to employ ``policy features'' learned from raw observations (e.g., raw images) of collaborators and opponents by inferring their policies. DPIQN incorporates the learned policy features as a hidden vector into its own deep Q-network (DQN), such that it is able to predict better Q values for the controllable agents than the state-of-the-art deep reinforcement learning models. We further propose an enhanced version of DPIQN, called deep recurrent policy inference Q-network (DRPIQN), for handling partial observability. Both DPIQN and DRPIQN are trained by an adaptive training procedure, which adjusts the network's attention to learn the policy features and its own Q-values at different phases of the training process. We present a comprehensive analysis of DPIQN and DRPIQN, and highlight their effectiveness and generalizability in various multi-agent settings. Our models are evaluated in a classic soccer game involving both competitive and collaborative scenarios. Experimental results performed on 1 vs. 1 and 2 vs. 2 games show that DPIQN and DRPIQN demonstrate superior performance to the baseline DQN and deep recurrent Q-network (DRQN) models. We also explore scenarios in which collaborators or opponents dynamically change their policies, and show that DPIQN and DRPIQN do lead to better overall performance in terms of stability and mean scores.
\end{abstract}

% AAMAS: the ACM CCS are encouraged but optional within AAMAS papers
%%
%% The code below should be generated by the tool at
%% http://dl.acm.org/ccs.cfm
%\begin{CCSXML}
%<ccs2012>
%  <concept_id>10010520.10010553.10010562</concept_id>
%  <concept_desc>Computer systems organization~Embedded systems</concept_desc>
%  <concept_significance>500</concept_significance>
%  <concept_id>10010520.10010575.10010755</concept_id>
%  <concept_desc>Computer systems organization~Redundancy</concept_desc>
%  <concept_significance>300</concept_significance>
%  <concept_id>10010520.10010553.10010554</concept_id>
%  <concept_desc>Computer systems organization~Robotics</concept_desc>
%  <concept_significance>100</concept_significance>
%  <concept_id>10003033.10003083.10003095</concept_id>
%  <concept_desc>Networks~Network reliability</concept_desc>
%  <concept_significance>100</concept_significance>
%</ccs2012>  
%\end{CCSXML}
%
%\ccsdesc[500]{Computer systems organization~Embedded systems}

\keywords{Deep Reinforcement Learning; Opponent Modeling; Multi-agent Learning}  % put your semicolon-separated keywords here!

\maketitle

%%%%%%%%%%%%%%%%%%%%%%%%%%%%%%%%%%%%%%%%%%%%%%%%%%%%%%%%%%%%%%%%%%%%%%%%%%%%%%%%%%%%%%%%%%%%%%%%%%%%%%%%%
%% start of main body of paper

\section{Introduction}\label{sec.Intro}
\noindent Modeling and exploiting other agents' behaviors in a multi-agent system (MAS) have received much attention in the past decade~\cite{lowe2017multi,he2016opponent,collins2007combining,littman1994markov}.  In such a system, agents share a common environment, where they can act and interact independently in order to achieve their own objectives.  The environment perceived by each agent, however, changes over time due to the actions exerted by the others, causing non-stationarity in each agent's observations.  A non-stationary environment prohibits an agent from assuming that the others have specific strategies and are stationary, leading to increased complexity and difficulty in modeling their behaviors.  In collaborative or competitive scenarios, in which agents are required to cooperate or take actions against the others, modeling environmental dynamics becomes even more challenging.   In order to act optimally under such scenarios, an agent needs to predict other agents' policies and infer their intentions. For scenarios in which the other agents' policies dynamically change over time, an agent's policy also needs to change accordingly. This further necessitates a robust methodology to model collaborators or opponents in an MAS. \\
\indent There is a significant body of work on MAS.  The literature contains numerous studies of MAS modeling~\cite{castaneda2016deep,bloembergen2015evolutionary,nowe2012game} and investigations of non-stationarity issues~\cite{hernandez2017survey}.  Most of previous researches on opponent or collaborator modeling, however, were domain-specific.  Such models either assume rule-based agents with substantial knowledge of environments~\cite{bai2015online,akiyama2012online}, or exclusively focus on one type of applications such as poker and real-time strategy games~\cite{schadd2007opponent,billings1998opponent}.  A number of early RL-based multi-agent algorithms have been proposed~\cite{banerjee2007generalized,conitzer2007awesome,bowling2004convergence,hu2003nash,bowling2002multiagent,singh2000nash}.   Some researchers presumed that the agent possesses a priori knowledge of some portions of the environment to ensure convergence~\cite{bowling2004convergence}.   Techniques presumed that the agent knows the underlying MAS structures~\cite{banerjee2007generalized,conitzer2007awesome,bowling2002multiagent} have been explored.  The use of other agents' actions or received rewards were suggested in \cite{conitzer2007awesome,hu2003nash}.  These assumptions are unlikely to hold in practical scenarios where an agent has no access to such information. While approaches for eliminating the need of prior knowledge of the environment have been attempted \cite{mealing2013opponent,zhang2010multiagent,abdallah2008multiagent,littman1994markov}, they are still limited to simple grid-world settings, and are unable to be scaled to more complex environments.\\
\indent In recent years, a special field called deep reinforcement learning (DRL), which combines RL and deep neural networks (DNNs), has shown great successes in a wide variety of single-agent stationary settings, including Atari games \cite{mnih2015human,hausknecht2015deep}, robot navigation \cite{zhang2016learning}, and Go \cite{silver2016mastering}. These advances lead researchers to start extending DRL to the multi-agent domain, such as investigating multi-agents' social behaviors~\cite{leibo2017multi,tampuu2015multiagent} and developing algorithms for improving the training efficiency~\cite{foerster2017stabilising,lowe2017multi,he2016opponent}. Recently, representation learning in the form of auxiliary tasks has been employed in several DRL methods~\cite{pathakICM17curiosity,jaderberg2016reinforce,mirowski2016learning,shelhamer2016loss}.  Auxiliary tasks are combined with DRL by learning additional goals \cite{jaderberg2016reinforce,mirowski2016learning,shelhamer2016loss}.
As auxiliary tasks provide DRL agents much richer feature representations than traditional methods, they are potentially more suitable for modeling non-stationary collaborators and opponents in an MAS. \\
\indent In the light of the above issues, we first present a detailed design of the deep policy inference Q-network (DPIQN), which aims at training and controlling a single agent to interact with the other agents in an MAS, using only high-dimensional raw observations (e.g., images).  DPIQN is built on top of the famous deep Q-network (DQN)~\cite{mnih2015human}, and consists of three major parts: a feature extraction module, a Q-value learning module, and an auxiliary policy feature learning module.  The former two modules are responsible for learning the Q values, while the latter module focuses on learning a hidden representation from the other agents' policies.  We call the learned hidden representation "policy features", and propose to incorporate them into the Q-value learning module to derive better Q values.  We further propose an enhanced version of DPIQN, called deep recurrent policy inference Q-network (DRPIQN), for handling partial observability resulting from the difficulty to directly deduce or infer the other agents' intentions from only a few observations~\cite{hausknecht2015deep}. DRPIQN differs from DPIQN in that it incorporates additional recurrent units into the DPIQN model. Both DPIQN and DRPIQN encourage an agent to exploit idiosyncrasies of its opponents or collaborators, and assume that no priori domain knowledge is given. It should be noted that in the most related work~\cite{he2016opponent}, the authors trained their agent to play a two-player game using handcrafted features and fixed the opponent's policy in an episode, which is not practical in real world environments. \\
\indent To demonstrate the effectiveness and generalizability of DPIQN and DRPIQN, we evaluate our models on a classic soccer game environment~\cite{collins2007combining,uther1997adversarial,littman1994markov}.  We jointly train the Q-value learning module and the auxiliary policy feature learning module at the same time, rather than separately training them with domain-specific knowledge.  Both DPIQN and DRPIQN are trained by an adaptive training procedure, which adjusts the network's attention to learn the policy features and its own Q-values at different phases of the training process.   We present experimental results in two representative scenarios: 1 vs. 1 and 2 vs. 2 games.  We show that DPIQN and DRPIQN are much superior to the baseline DRL models in various settings, and are scalable to larger and more complex environments.  We further demonstrate that our models are generalizable to environments with unfamiliar collaborator or opponents.  Both DPIQN and DRPIQN are able to change their strategies in response to the other agents' moves.

The contributions of this work are as follows:
\vspace{-0.3em}
\begin{itemize}
\itemsep-0.1em 
\item DPIQN and DRPIQN enable an agent to collaborate or compete with the others in an MAS by using only high-dimensional raw observations. 

\item DPIQN and DRPIQN incorporate policy features into their Q-value learning module, allowing them to derive better Q values in an MAS than the other DRL models.

\item An adaptive loss function is used to stabilize the learning curves of DPIQN and DRPIQN.

\item Unlike the previous works~\cite{he2016opponent,billings1998opponent,uther1997adversarial} only focusing on competitive environments, our models are capable of handling both competitive and collaborative environments.

\item Our models are generalizable to unfamiliar collaborators or opponents.
\end{itemize}
\vspace{-0.3em}
The remainder of this paper is organized as follows.  Section 2 introduces background materials related to this paper.  Section 3 describes the proposed DPIQN and DRPIQN models, as well as the training and generalization methodologies.  Section 4 presents our experimental results, and provides a comprehensive analysis and evaluation of our models. Section 5 concludes this paper.

\section{Background}\label{sec.background}
\noindent RL is a technique for an agent to learn which action to take in each of the possible states of an environment $\mathcal{E}$.   The goal of the agent is to maximize its accumulated long-term rewards over discrete time steps \cite{Sutton:1998:IRL:551283,littman1994markov}.  The environment $\mathcal{E}$ is usually formulated as a Markov decision process (MDP), represented as a 5-tuple $(s, a, \mathcal{T}, \mathcal{R}, \gamma)$.  At each timestep, the agent observes a state $s \in \mathcal{S}$, where $\mathcal{S}$ is the state space of $\mathcal{E}$.  It then performs an action $a$ from the action space $\mathcal{A}$, receives a real-valued scalar reward $r$ from $\mathcal{E}$, and moves to the next state $s^{\prime} \in \mathcal{S}$.  The agent's behavior is defined by a policy $\pi$, which specifies the selection probabilities over actions for each state.  
The reward $r$ and the next state $s^{\prime}$ can be derived by $r = \mathcal{R}(s, a, s^\prime)$ and $\mathcal{T}(s^\prime, s, a) = \text{Pr}(s^\prime | s, a)$,
where $\mathcal{R}$ and $\mathcal{T}$ are the reward function and the transition probability function, respectively.  
Both $\mathcal{R}$ and $\mathcal{T}$ are determined by $\mathcal{E}$.
The goal of the RL agent is to find a policy $\pi$ which maximizes the expected return $G_{t}$, which is the discounted sum of rewards given by $G_t = \sum^{T}_{\tau = t}\gamma^{\tau-t}r_{\tau}$,
where $T$ is the timestep when an episode ends, $t$ denotes the current timestep, $\gamma \in [0, 1]$ is the discount factor, and $r_{\tau}$ is the reward received at timestep $\tau$.  The action-value function (abbreviated as Q-function) of a given policy $\pi$ is defined as the expected return starting from a state-action pair $(s, a)$, expressed as $Q^{\pi}(s,a) = \mathbb{E}\big[G_{t}|s_{t} = s, a_{t} = a, \pi]$.

The optimal Q-function $Q^{*}(s, a)$, which provides the maximum action values for all states, is determined by the \textit{Bellman optimality equation} \cite{Sutton:1998:IRL:551283}:
\vspace{-0.3em}
\begin{equation}\label{eq::bellmaneq}
	Q^{*}(s,a) = \sum_{s^\prime}{\mathcal{T}}(s^\prime, s, a)\big[r + \gamma\max_{a^\prime} Q^{*}(s^\prime,a^\prime)\big]
\vspace{-0.3em}
\end{equation}
where $a^\prime$ is the action to be selected in state $s^\prime$.
An optimal policy $\pi^{*}$ is then derived from Eq.~\eqref{eq::bellmaneq} by selecting the highest-valued action in each state, and can be expressed as $\pi^{*}(s) = \argmax_{a \in \mathcal{A}}Q^{*}(s, a)$.

\subsection{Deep Q-Network}\label{sec.background.dqn}
DQN \cite{mnih2015human} is a model-free approach to RL based on DNNs for estimating the Q-function over high-dimensional and complex state space. DQN is parameterized by a set of network weights $\theta$, which can be updated by a variety of RL algorithms~\cite{mnih2015human,hausknecht2015deep}. 
To approximate the optimal Q-function given a policy $\pi$ and state-action pairs $(s,a)$, DQN incrementally updates its set of parameters $\theta$ such that $Q^{*}(s,a) \approx Q(s,a, \theta)$.

The parameters $\theta$ are learned by gradient descent which iteratively minimizes the loss function $L(\theta)$ using samples $(s, a, r, s^{\prime})$ drawn from an experience replay memory $Z$.  $L(\theta)$ is expressed as:
\begin{equation} \label{eq::q_loss}
	L(\theta) = \mathbb{E}_{s,a,r,s^\prime \sim U(Z)}\big[(y - Q(s,a, \theta))^2\big]
\end{equation}
where $y = r + \gamma\max_{a^\prime} Q(s^\prime,a^\prime, \theta^{-})$, $U(Z)$ is a uniform distribution over $Z$, and $\theta^{-}$ represents the parameters of the target network.   The target network is the same as the online network, except that its parameters $\theta^{-}$ are updated by the online network at predefined intervals.  Both the experience replay memory and the target network enhance stability of the learning process dramatically.

\subsection{Deep Recurrent Q-Network}\label{sec.background.drqn}
Deep recurrent Q-network (DRQN) is proposed by~\cite{hausknecht2015deep} to deal with partial observability caused by incomplete and noisy state information in real-world tasks. It is developed to train an agent in an environment modeled as a partial observable Markov decision process (POMDP), in which the state of the environment is not fully observable or determinable from a limited number of past states.  DRQN models $\mathcal{E}$ as a 6-tuple $(s, a, \mathcal{T}, \mathcal{R}, \gamma, o)$, where $o$ is the observation perceived by the agent.  Instead of using only the last few states to predict the next action as DQN, DRQN extends the architecture of DQN with Long Short Term Memory (LSTM)~\cite{hochreiter1997long}. It integrates the information across observations by an LSTM layer to a hidden state $h$ and an internal cell state $c$,
which recurrently encodes the information of the past observations.  The Q-function of DRQN is represented as $Q(o, h, a)$.  DRQN has been demonstrated to perform better than DQN at all levels of partial information~\cite{hausknecht2015deep}.
\begin{figure}[!tb]
	\centering
    % width = 0.675 before
    \includegraphics[width=0.6\linewidth]{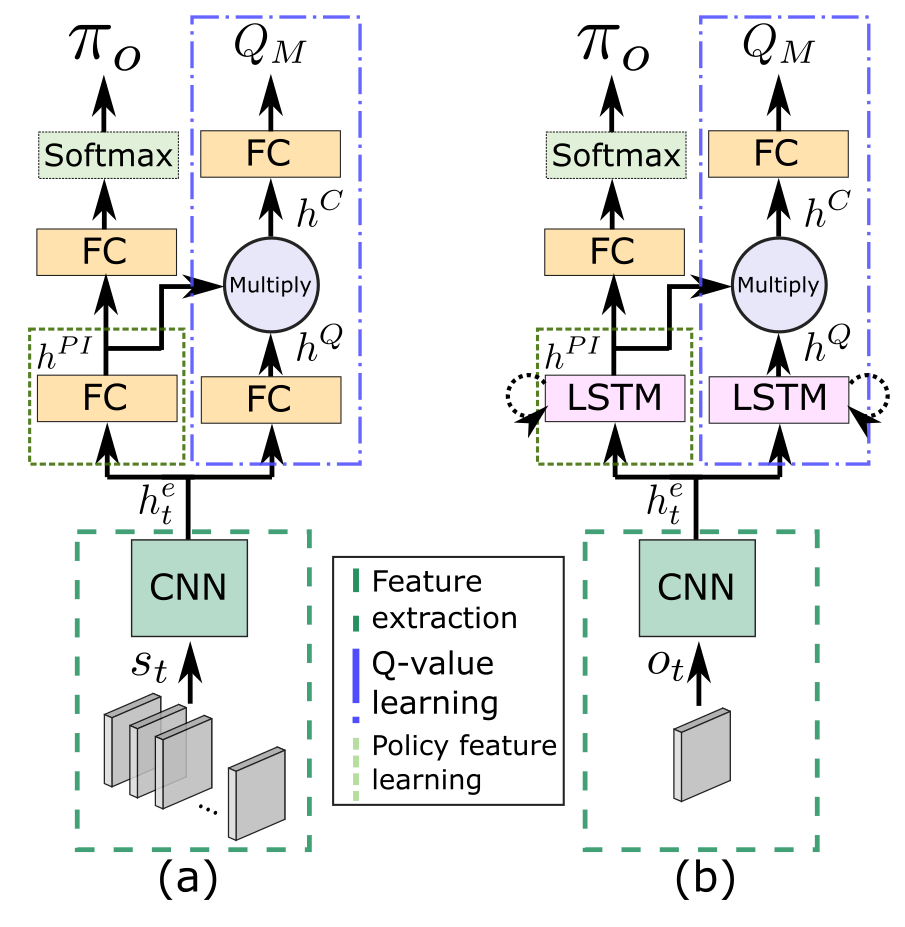}
    \caption{Architectures of DPIQN (a) and DRPIQN (b).}
    \label{figure::model_archi}
    \vspace{-2.5em}
\end{figure}
\subsection{Q-learning in Multi-Agent Environments}
\noindent In a real-world MAS, the  environment state is affected by the joint action of all agents.  This means that the state perceived by each agent is no longer stationary.  The Q-function of an agent is thus dependent on the actions of the others.  Adaptation of the Q-function definition becomes necessary, such that the other agents' actions are taken into consideration.

In order to re-formulate the Q-function for multi-agent settings, we assume that the environment $\mathcal{E}$ contains a group of $N+1$ agents: a controllable agent and the other $N$ agents.  The latter can be either collaborators or opponents.   The joint action of those $N$ agents is defined as $a_o \in \mathcal{A}_o$, where $\mathcal{A}_o \colon \mathcal{A}_1 \bigtimes \mathcal{A}_2 \cdots \bigtimes \mathcal{A}_N$, and $\mathcal{A}_1$$\sim$$\mathcal{A}_N$ are the action spaces of agents $1 \sim N$, respectively.
The reward function $\mathcal{R}_M$ and the state transition function $\mathcal{T}_M$, therefore, become $r = \mathcal{R}_M(s, a, a_o, s^\prime)$ and $\mathcal{T}_M(s^\prime, s, a, a_o) = \text{Pr}(s^\prime | s, a, a_o)$, where the subscript $M$ denotes multi-agent settings. 
We further define the other $N$ agents' joint policy as $\pi_o(a_o|s)$.  Based on 
$\mathcal{R}_M$ and $\mathcal{T}_M$, Eq.~\eqref{eq::bellmaneq} can then be rewritten as:
\begin{equation} \label{eq::ma_bellmaneq}
	\begin{split}
		{Q_M}^*(s, a | \pi_o) = \sum_{a_o}{\pi_o(a_o|s)}\sum_{s^\prime}{\mathcal{T}_M}(s^\prime, s, a, a_o) \\ \big[\mathcal{R}_M(s, a, a_o, s^\prime) + \gamma\mathbb{E}_{a^\prime}\big[{Q_M}^{*}(s^\prime,a^\prime|\pi_o)\big]\big]
	\end{split}
\end{equation}
Please note that in the above equation, the Q-function is conditioned on $\pi_o$, rather than $a_o$.  Conditioning the Q-function on $a_o$ leads to an explosion of the number of parameters, and is not suitable for an MAS with multiple agents.

% figure for sec 3
\begin{figure}[!tb]
    % width = 0.675 before
    \includegraphics[width=0.6\linewidth]{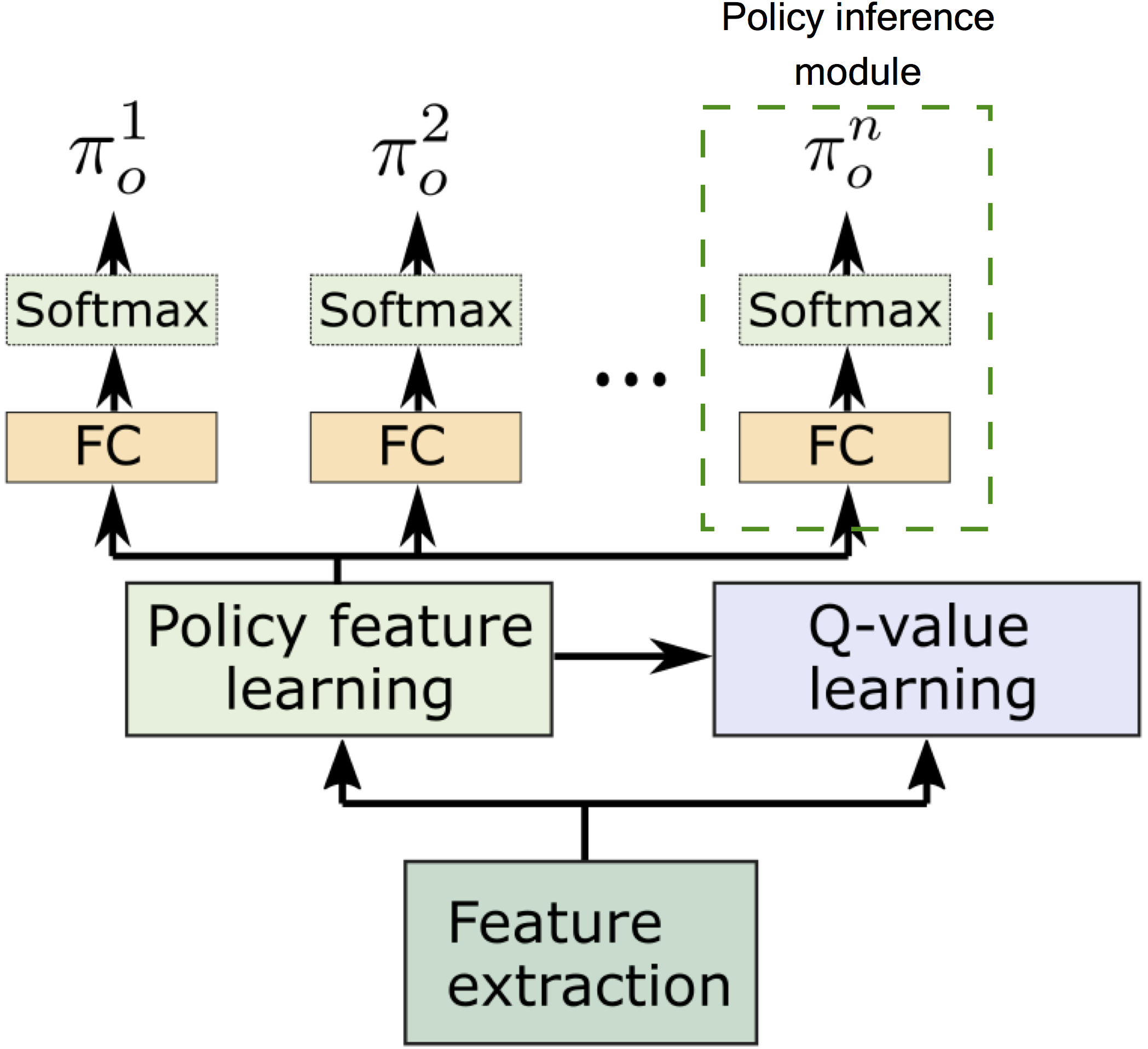}
	\caption{Generalized architecture of DPIQN and DRPIQN.}
    \label{fig::generalization}
    \vspace{-1.5em}
\end{figure}

% The figures for sec 3 is moved to background due to formatting issues
\section{Deep Policy Inference Q-Network}\label{sec.dpiqn}
%\begin{figure}[!tb]
%	\centering
%    \begin{minipage}[b]{0.49\linewidth}
%    	\includegraphics[width=\linewidth]{model_diagram.png}
%        \caption{The architextures of DPIQN (a) and DRPIQN (b).}
%        \label{figure::model_archi}
%    \end{minipage}
%    \hfill
%    \begin{minipage}[b]{0.49\linewidth}
%    	\includegraphics[width=\linewidth]{generalization.png}
%        \caption{The generalized architecture of DPIQN and DRPIQN.}
%        \label{fig::generalization}
%    \end{minipage}
%\end{figure}
\noindent In this section, we present the architecture and implementation details of DPIQN. We first outline the network structure, and introduce the concept of policy features.  Then, we present a variant version of DPIQN, called DRPIQN, for dealing with partial observability.  We next explain the training methodology in detail.  Finally, we discuss the generalization methodology for environments with multiple agents.
\subsection{DPIQN}
\noindent 
The main objective of DPIQN is to improve the quality of state feature representations of an agent in multi-agent settings.  As mentioned in the previous section, the environmental dynamics are affected by multiple agents.   In order to enhance the hidden representations such that the controllable agent can exploit the other agents' actions in an MAS, DPIQN learns the other agents' ``policy features'' by auxiliary tasks.   Assume that the environment contains a controllable agent and a target agent.  Policy features are defined as a hidden representation used by the controllable agent to infer the policy of the target agent, and are represented as a vector $h^{PI}$.   The approximated representation encodes the spatial-temporal features of the target agent's policy $\pi_o$, and can be obtained by observing the target agent's behavior for a series of consecutive timesteps.  By incorporating $h^{PI}$ into the model of the controllable agent, DPIQN is able to learn a better Q-function than traditional methodologies.  DPIQN adapts itself to dynamic environments in which the target agent's policy changes over time.  The results are presented in Section 4. \\
\indent \sloppy The network architecture of DPIQN is illustrated in Fig. \ref{figure::model_archi} (a), which computes the controllable agent's Q-value $Q_M(s,a|h^{PI},h^e_t;\theta)$ and the target agent's policy $\pi_{o}$.  Note that $Q_M$ is conditioned on the policy feature vector $h^{PI}$, rather than the target agent's action $a_o$ or policy $\pi_o$. We omit $h^{PI}$ and $h^e_t$ in $Q_M$ in the rest of the paper for simplicity.  DPIQN consists of three parts: a feature extraction module, a Q-value learning module, and an auxiliary policy feature learning module. The feature extraction module is a convolutional neural network (CNN) shared between the latter two modules, and is responsible for extracting the spatio-temporal features from the latest $k$ inputs (i.e. observations). The extracted feature at timestep $t$ are denoted as $h^e_t$. In our experiments presented in Section 4, the inputs applied at the feature extraction module are raw pixel data.  \\
\indent Both the Q-value learning module and the policy feature learning module take $h^e_t$ as their common input.  In DPIQN, these two modules are composed of a series of fully-connected layers (abbreviated as FC layers).  The Q-value learning module is trained to approximate the optimal Q-function $Q^*_M$, while the policy feature learning module is trained to infer the target agent's next action $a_o$.  Based on $h^e_t$, these two modules separately extract two types of features denoted as $h^Q$ and $h^{PI}$ using two FC layers followed by non-linear activation functions.  The policy feature vector $h^{PI}$ is further fed into an FC layer and a softmax layer, which are collectively referred to as a policy inference module, to generate the target agent's approximated policy $\pi_o (a_o | h^e_t)$.
The approximated policy $\pi_o$ is then used to compute the cross entropy loss against the true action of the target agent.  The training procedure is discussed in subsection 3.3.
To derive the Q values of the controllable agent, the policy feature vector $h^{PI}$ is fed into the Q-value learning module, and merged with $h^Q$ by multiplication to generate $h^C$, as annotated in Fig. \ref{figure::model_archi}. $h^C$ is then processed by an FC layer to generate the Q-value of the controllable agent.
\subsection{DRPIQN}
\noindent 
DRPIQN is a variant of DPIQN motivated by DRQN for handling partial observability, with an emphasis on decreasing the hidden state representation noise from strategy changing of the other agents in the environment. For example, the policy $\pi_o$ of an opponent in a competitive task may switch from a defensive mode to an offensive mode in an episode, leading to an increased difficulty in adapting the Q-function and the approximated policy feature vector $h^{PI}$ of the controllable agent to such variations. This becomes even more severe in multi-agent settings when the policies of all agents change over time, resulting in degradation in the stability of $h^{PI}$ (and hence, $h^C$).  In such environments, inferring the policy of a target agent becomes a POMDP problem: the intention of the target agent cannot be directly deduced or inferred from only a few observations.

DRPIQN is proposed to incorporate recurrent units in the baseline DPIQN model to deal with the above issues, as illustrated in Fig. \ref{figure::model_archi} (b). DRPIQN takes a single observation as its input. It similarly employs a CNN to extract spatial features $h^e$ from the input, but uses the LSTM layers to encode temporal correlations between a history of them. Due to its capability of learning long-term dependencies, the LSTM layers are able to capture a better policy feature representation. We show in Section 4 that DRPIQN demonstrates superior generalizability to unfamiliar agents than the baseline models.

\subsection{Training with Adaptive Loss}
\noindent In this section, we provide an overview of the training methodology used for DPIQN and DRPIQN. Algorithm~\ref{sec.dqnpi.training} provides the pseudocode of the training procedure.  Our training methodology stems from that of DQN, with a modification of the definition of loss function.  We propose to adopt two different loss function terms $L^Q$ and $L^{PI}$ to train our models.  The former is the standard DQN loss function.
The latter is called the policy inference loss, and is obtained by computing the cross entropy loss between the inferred policy $\pi_o$ and the ground truth one-hot action vector $\mu_o$ of the target agent. $L^{PI}$ is expressed as:
\vspace{-0.3em}
\begin{equation}
\begin{split}
L^{PI} &=  H(\mu_{o}) + D_{KL}(\mu_{o}\|\pi_{o})
\end{split}
\vspace{-0.3em}
\end{equation}
where $H(\mu_o)$ is the entropy of $\mu_o$, and $D_{KL}(\mu_o || \pi_o)$ stands for the Kullback-Leibler divergence of $\pi_o$ from $\mu_o$.
The aggregated loss function can be expressed as:
\vspace{-0.3em}
\begin{equation} \label{eq::dpiqn_loss}
L = E_{mini\text{-}batch \sim U(Z)} [(\lambda L^Q + L^{PI})]
\vspace{-0.3em}
\end{equation}
where $\lambda$ is called the adaptive scale factor of $L^Q$. The function of $\lambda$ is to adaptively scale $L^Q$ at different phases of the training process. It is defined as: 

\begin{equation} \label{eq::lambda}
	\lambda = \frac{1}{\sqrt[]{L^{PI}_t}}
\end{equation}
In the initial phase of the training process, $L^{PI}$ is large, corresponding to a small $\lambda L^Q$. A small $\lambda L^Q$ encourages the network to focus on learning the policy feature vector $h^{PI}$. When the network is trained such that $L^{PI}$ is sufficiently small, $\lambda L^Q$ becomes dominant in Eq.~\eqref{eq::dpiqn_loss}, turning the network's attention to optimize the Q values.  We found that without the use of $\lambda$, the Q values tend to converge in an optimistic fashion, leading to a degradation in performance.  The intuition behind $\lambda$ is that if the controllable agent possesses sufficient knowledge of the target agent, it is able to exploit this knowledge to make a better decision.  The use of $\lambda$ significantly improves the stability of the learning curve of the controllable agent.  A comparison of performance with and without the usage of $\lambda$ is presented in Section 4.5.  Note that during testing, the forward path only calculates the Q values of the controllable agent, and does not need to be fed with the moves of the target agent.

\begin{algorithm}[tb]
\caption{Training Procedure of DPIQN}\label{sec.dqnpi.training}
\begin{algorithmic}[1]
\State Initialize replay memory $Z$, environment $\mathcal{E}$, and observation $s$
\State Initialize network weights $\theta$
\State Initialize target network weights $\theta^{-}$
\For{timestep $t=1$ to $T$}
  \State Take $a$ with $\epsilon$-greedy based on $Q_M(s,a;\theta)$
  \State Execute $a$ in $\mathcal{E}$ and observe $\mu_{o}$, $r$, $s^{\prime}$
  \State Clip $r$ between $[-1,1]$
  \State Store transition $(s,a,a_{o},r,s^{\prime})$ in $Z$
  \State Sample mini-batch $(s^j,a^j,a^j_{o},r^j,s^{j\prime})$ from $Z$ 
  \State Set~$y = \begin{cases}
  	r^j & \text{for terminal~} s^{j\prime} \\
    r^j+\gamma\max\limits_{a^{j\prime}}Q_M(s^{j\prime},a^{j\prime};\theta^-) & \text{otherwise}
  	\end{cases}$
  \State Compute $L^Q$ based on $y$, and compute $L^{PI}$
  \State Compute $\lambda=\frac{1}{\sqrt[]{L^{PI}}}$
  \State Compute gradient $G$ based on $L^{Q}$, $L^{PI}$, $\lambda$
  \State Perform gradient descent $G$
  \State Update $s=s^{\prime}$ 
  \State Update $\theta^-=\theta$ for every $C$ steps
\EndFor
\State \textbf{end}
\end{algorithmic}
\end{algorithm}
% figures from sec 4
% The scales in the soccer illustration
\newcommand\scaleSoccer{0.6}
\newcommand\scaleLegend{0.3}
\newcommand\scalePlayer{0.5}
% The field width and height
\newcommand\fx{9}
\newcommand\fy{6}
% The commands to draw the overlays
\newcommand\DrawImage[4]{
  % The boundary node
  \node[fit={($ (1.0/\fx*#2,1.0/\fy*#3) $) ($ ({1.0/\fx*(#2+1)},{1.0/\fy*(#3+1)}) $)},inner sep=0] (#1) {};
  % The image at the center of the boundary node
  \node[anchor=center,inner sep=0] at (#1.center) {\includegraphics[scale=\scalePlayer]{#4}};
}
\newcommand\DrawArrow[3]{
  \draw[->,thick,black] (#1) -- ([shift={(1.0/\fx*#2,1.0/\fy*#3)}]#1);
}
\newcommand\DrawText[5]{
  % The boundary node
  \node[fit={($ (1.0/\fx*#2,1.0/\fy*#3) $) ($ ({1.0/\fx*(#2+1)},{1.0/\fy*(#3+1)}) $)},inner sep=0] (#1) {};
  % The text at the center of the boundary node
  \node[scale=0.6,text=black] at (#1.center) {$ \pmb{#4} $};
}
\newcommand\DrawRect[5]{
  \node[fit={($ (1.0/\fx*#2,1.0/\fy*#3) $) ($ (1.0/\fx*#4,1.0/\fy*#5) $)},inner sep=0pt,dashed,draw=red,very thick] (#1) {};
}

\begin{figure}[!t]
	% References:
    % https://tex.stackexchange.com/a/9561
    % https://tex.stackexchange.com/a/302973
    % https://tex.stackexchange.com/a/197412
    \begin{tikzpicture}
    % Background image
    \node[anchor=north west,inner sep=0] (background) at (0,0) {\includegraphics[width=\scaleSoccer\linewidth]{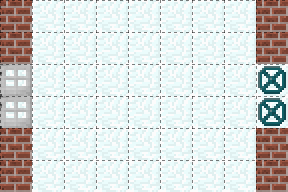}};
    % Overlays
    \begin{scope}[x={(background.north east)},y={(background.south west)}]
      \DrawImage{s1_p1}{3}{2}{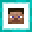}
      \DrawArrow{s1_p1.east}{0.5}{0};
      \DrawArrow{s1_p1.north}{0}{-0.5};
      \DrawArrow{s1_p1.west}{-0.5}{0};
      \DrawArrow{s1_p1.south}{0}{0.5};
      \DrawImage{s1_p2}{5}{3}{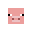}
      \DrawArrow{s1_p2.east}{0.5}{0};
      \DrawArrow{s1_p2.north}{0}{-0.5};
      \DrawArrow{s1_p2.west}{-0.5}{0};
      \DrawArrow{s1_p2.south}{0}{0.5};
      \DrawText{s1_label1}{2.2}{1.3}{(1)};
      \DrawText{s1_label2}{5.8}{2.3}{(2)};
      \DrawRect{spawn_player}{1.15}{0.15}{3.85}{5.85};
      \DrawRect{spawn_opponent}{5.15}{0.15}{7.85}{5.85};
      \DrawText{spawn_label}{4.0}{0.0}{(3)};
    \end{scope}
    % Legend image
    \node[anchor=west,inner sep=0,right=1bp of background] (background) {\includegraphics[width=\scaleLegend\linewidth]{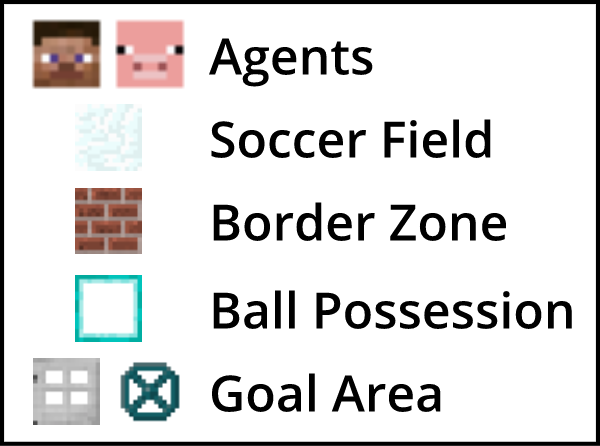}};
    \end{tikzpicture}
}{
	\captionsetup{hypcap=false}
	\caption{Illustration of the soccer game in 1 vs. 1 scenario. (1) is our controllable agent, (2) is the rule-based opponent, (3) is the start zone of the agents. The agent who possesses the ball is surrounded by a blue square.}
 	\label{figure::soccer_illustration}
\end{figure}
\begin{figure}[!t]
  \includegraphics[width=.8\linewidth]{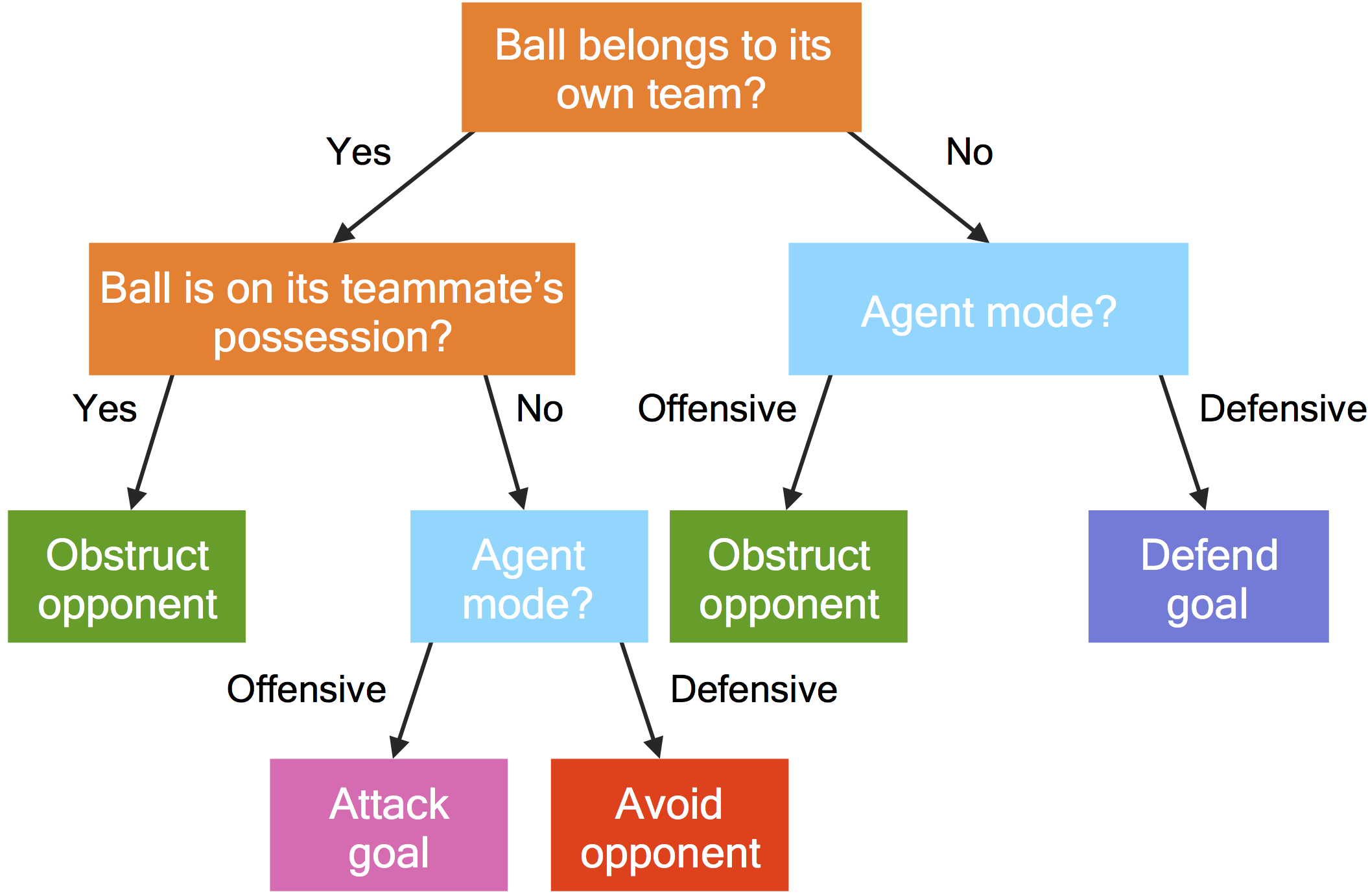}
  \caption{Policy of the rule-based agent in each episode.}
  \label{figure::policy}
  \vspace{-1.5em}
\end{figure}

\subsection{Generalization}
\noindent DPIQN and DRPIQN are both generalizable to complex environments with multiple agents.  Consider a multi-agent environment in which both cooperative and competitive agents coexist (e.g., a soccer game).  The policies of these agents are diverse in terms of their objectives and tactics.  The aim of the collaborative agents (collaborators) is to work with the controllable agent to achieve a common goal, while that of the competitive agents (opponents) is to act against the controllable agent's team. Some of the agents are more offensive, while others are more defensive.  In such a heterogeneous environment, conditioning the Q-function of the controllable agent on the actions of distinct agents would lead to an explosion of parameters, as mentioned in Section 3.3.  To reduce model complexity and concentrate the controllable agent's focus on the big picture, we propose to summarize the policies of the other agents in a single policy feature vector $h^{PI}$.  We extend the policy feature learning module in DPIQN and DRPIQN to incorporate multiple policy inference modules to learn the other agents' policies separately, as illustrated in Fig.~\ref{fig::generalization}.  Each policy inference module corresponds to either a collaborator or an opponent.  The loss function term $L^{PI}$ in Eq.~\eqref{eq::dpiqn_loss} is then modified as:
\begin{equation}
\begin{split}
L^{PI} &=  \frac{1}{N} \sum^N_{i = 0} H(\mu^i_o) + D_{KL}(\mu^i_o\|\pi^i_o)
\end{split}
\end{equation}
where $N$ is the total number of the target agents, and $i$ indicates the $i$-th target agent.  The training procedure is the same as Algorithm~\ref{sec.dqnpi.training}, except lines 8$\sim$10 are modified to incorporate $a^i_o$, $\pi^i_o$, and $\mu^i_o$.  The learned $h^{PI}$, therefore, embraces the policy features from the collaborators and opponents.  The experimental results of the proposed generalization scheme is presented in Sections 4.3, 4.4 and 4.5.

% Some figures are moved to sec 3 for formatting issues
\section{Experimental Results} \label{sec.experiment}
\noindent In this section, we present experimental results and discuss their implications.  We start by a brief introduction to our experimental setup, as well as the environment we used to evaluate our models. 

\begin{table}[!t]
	\caption{Hyperparameters used in our experiments.}
    \label{table::hyperparam}
	\renewcommand{\arraystretch}{1.65}
	\centering
    \small
    \begin{tabular}{l|cm{4cm}}
    \Xhline{2\arrayrulewidth}
    \multicolumn{2}{c}{List of Hyperparameters} 			\\ \hline \hline
    Epoch length  	  	& 10,000 timesteps (2,500 training steps)				\\ \hline
    Optimizer 	  	    & Adam~\cite{kingma2014adam}					\\ \hline
    Learning rate 	  	& \stackanchor{0.001}{\footnotesize (Initial)} $\rightarrow$ \stackanchor{0.0004}{\footnotesize (Epoch 600)} $\rightarrow$ \stackanchor{0.0002}{\footnotesize (Epoch 1000)} 	   \\ \hline
    $\epsilon$-greedy 	& \stackanchor{1.0}{\footnotesize (Initial)} $\rightarrow$ \stackanchor{0.1}{\footnotesize (Epoch 100)} 		   \\ \hline
    Discount factor   	& 0.99 							\\ \hline
    Target network 	    & Update every 10,000 steps 	\\ \hline
    Replay memory     	& 1,000,000 samples 			\\ \hline
    History length    	& \stackanchor{12}{\footnotesize (DQN/DPIQN)} \stackanchor{1}{\footnotesize (DRQN/DRPIQN)}							\\ \hline
    Minibatch size    	& 32 							\\
    \Xhline{2\arrayrulewidth}
    \end{tabular}
    %\vspace{0.5em}
\end{table}
\subsection{Experimental Setup}
\noindent We perform our experiments on a soccer game environment illustrated in Fig.~\ref{figure::soccer_illustration}.  
We begin with explaining the environments and game rules. The hyperparameters we used are summarized in Table~\ref{table::hyperparam}.
The source code is developed based on Tensorpack\footnote{github.com/ppwwyyxx/tensorpack}, which is a neural network training interface built on top of TensorFlow~\cite{abadi2016tensorflow}.

\begin{table}[!t]
\caption{Evaluation result of 1 vs. 1 scenario.}
\label{table::soccer_1vs1_reward}
\centering
\small
\begin{tabular}{l|c|C{3em}C{3.8em}C{3.8em}}
\Xhline{2\arrayrulewidth}
Model                 & Training  & Hybrid                     & Offensive                     & Defensive                     \\ \hline
                      & Hybrid    & -0.063                     & -0.850                        & 0.000                         \\
DQN                   & Offensive & 0.312                      & \underline{0.658}             & 0.113                         \\
                      & Defensive & -0.081                     & -1.000                        & \underline{0.959}             \\
DRQN                  & Hybrid    & 0.028                      & -0.025                        & 0.168                         \\ \hline \hline
DPIQN                 & Hybrid    & \textbf{0.999}             & \textbf{0.989}                & 0.986                         \\ 
DRPIQN                & Hybrid    & \textbf{0.999}             & 0.981                         & \textbf{1.000}                \\
\Xhline{2\arrayrulewidth}
\end{tabular}
%\vspace{0.5em}
\end{table}

\begin{table*}[tb]
	\centering
    \begin{minipage}[t]{0.3\textwidth}
    \vspace{0pt}
    	\includegraphics[width=\linewidth]{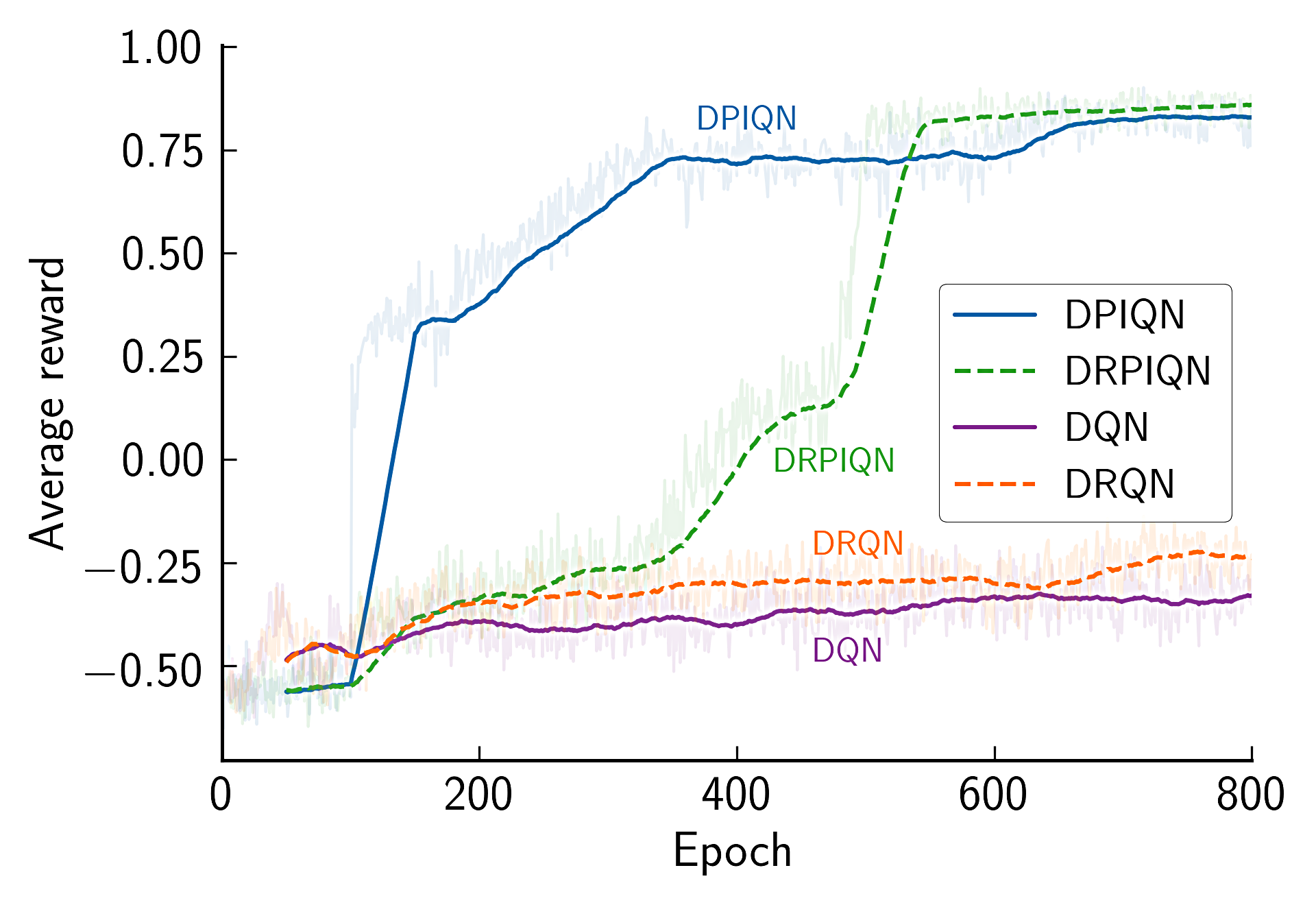}
        \captionof{figure}{Learning curve comparison in the 1 vs. 1 scenario.}
        \label{figure::learning_curves_1vs1}
    \end{minipage}%
    \hfill
    \begin{minipage}[t]{0.3\textwidth}
    \vspace{0pt}
    	\includegraphics[width=\linewidth]{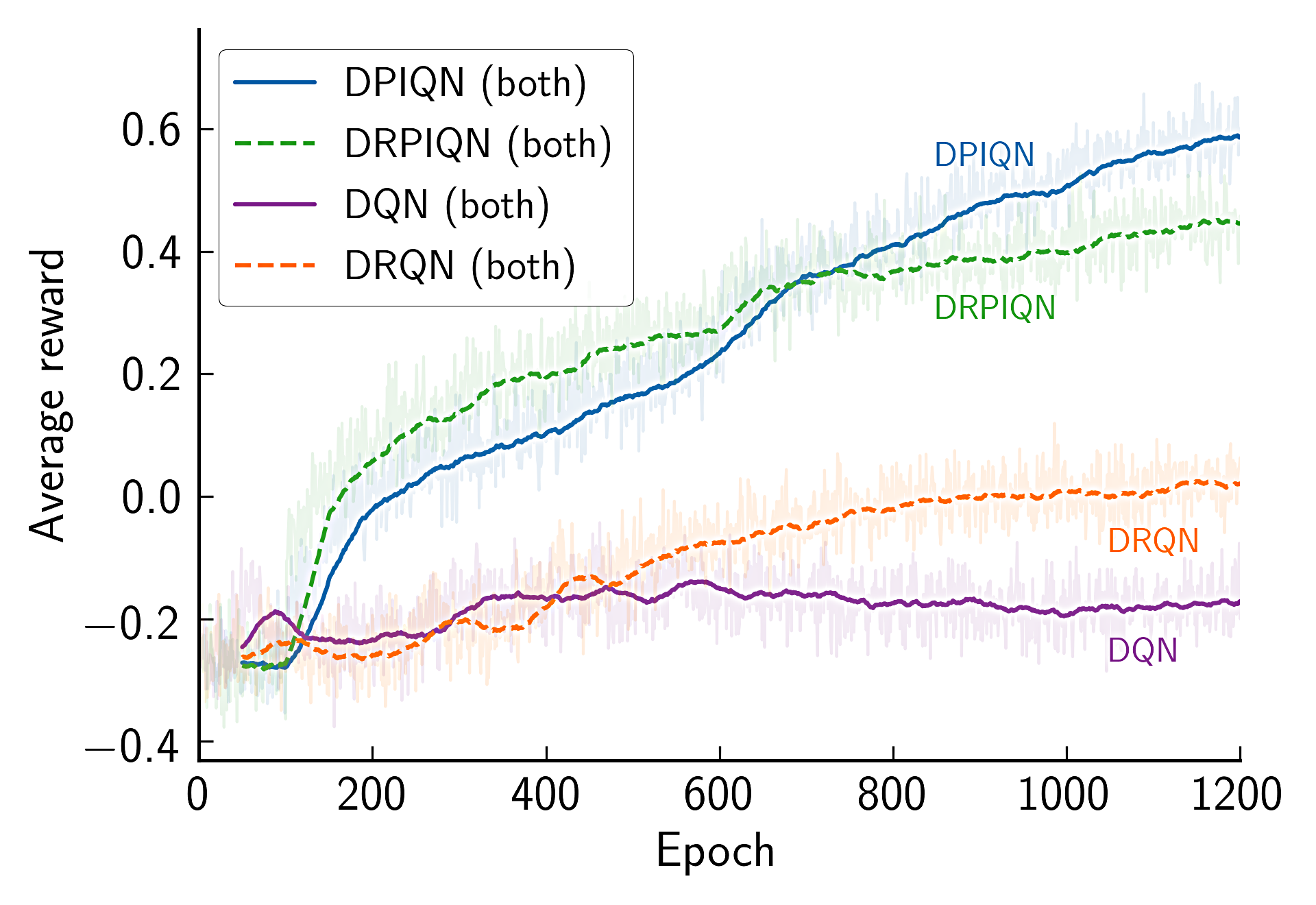}
        \captionof{figure}{Learning curve comparison in the 2 vs. 2 scenario.}
        \label{figure::learning_curves_2vs2}
    \end{minipage}%
    \hfill
    \begin{minipage}[t]{0.3\textwidth}
    \vspace{0pt}
    	\caption{Evaluation result of 2 vs. 2 scenario.}
		\label{table::soccer_2vs2_reward}
		\centering
		\begin{tabular}{l|c|C{3em}}
		\Xhline{2\arrayrulewidth}
		Model                      & Training & Testing                     \\ \hline
		DQN                        & -        & -0.152                      \\
		DRQN                       & -        & -0.031                      \\ \hline \hline
        		                   & Both     & 0.761                       \\
		DPIQN                      & O-only   & 0.645                       \\
                		           & C-only   & 0.233                       \\ \hline
                        		   & Both     & 0.695                       \\
		DRPIQN                     & O-only   & 0.714                       \\
        		                   & C-only   & \textbf{0.854}              \\          
		\Xhline{2\arrayrulewidth}
		\end{tabular}
    	%\includegraphics[width=\linewidth]{mean_reward_1vs1_reg_lamb.png}
        %\caption{Impact of $\lambda$ and $L^{PI}$ on average reward}
        %\label{figure::mean_reward_1vs1_reg_lamb}
    \end{minipage}%
    %\vspace{-1.5em}
\end{table*}

\paragraph{Environment.} Fig.~\ref{figure::soccer_illustration} illustrates the soccer game environment used in our experiments.  The soccer field is a grid world composed of multiple grids of 32$\times$32 RGB pixels and is divided into two halves.  The game starts with the controllable agent and the collaborator (Fig.~\ref{figure::soccer_illustration}(1)) randomly located on the left half of the field, and the opponents (Fig.~\ref{figure::soccer_illustration}(2)) randomly located on the right half of the field, except the goals and border zones (Fig.~\ref{figure::soccer_illustration}(3)). The initial possession of the ball (highlighted by a blue rectangle) and the modes of the agents (offensive or defensive) are randomly determined for each episode.  In each episode, each team's objective is to deliver the ball to the opposing team's goal.  An episode terminates immediately once a team scores a goal.  A reward of 1 is awarded if the controllable agent's team wins, while a penalty reward of -1 is given if it loses the match.  If neither of the teams is able to score within 100 timesteps, the episode ends with a reward of 0.  Each agent in the field chooses from five possible actions: \textit{move, N, S, W, E} and \textit{stand still} at each time step. If the intended position of an agent is out of bounds or overlaps with that of the other agents, the move doesn't take place.  In the latter case, the agent who originally possesses the ball passes/loses it to the collaborator/opponent who intend to move to the same position.  In our simulations, the controllable agent receives inputs in the form of resized grayscale images $o \in \mathbb{R}^{84\times 84\times 1}$ of the current entire world (Fig.~\ref{figure::soccer_illustration}).

\paragraph{1 vs. 1 Scenario.} In this scenario, the game is played by a controllable agent and an opponent on a $6 \times 9$ grid world (Fig.~\ref{figure::soccer_illustration}). The opponent is a two-mode rule-based agent playing according to Fig.~\ref{figure::policy}.  In the offensive mode, the opponent focuses on scoring a goal, or stealing the ball from the controllable agent.  In the defensive mode, the opponent concentrates on defending its own goal, or moving away from the controllable agent when it possesses the ball.
We set the frame skip rate to 1, which is optimal for DQN after an exhaustive search of hyperparameters.
\paragraph{2 vs. 2 Scenario.} In this scenario, each of the two teams contains two agents.  The two teams compete against each other on a grid world with larger areas of goals and border zones. We consider two tasks in this scenario:  our controllable agent has to collaborate with (1) a rule-based collaborator or (2) a learning agent to compete with the rule-based opponents. The sizes of the grid world are $13 \times 10$ and $21 \times 14$, respectively. In addition, the rule-based agents in this scenario also play according to Fig.~\ref{figure::policy}. When a teammate has the ball, the rule-based agent does its best to obstruct the nearest opponent. We set the frame skip rate to 2 in this scenario due to the increased size of the environment.

\subsection{Performance Comparison in 1 vs.1 Scenario}
\noindent Table~\ref{table::soccer_1vs1_reward} compares the controllable agent's average rewards among three types of the opponent agent's modes in the testing phase for four types of models, including DQN, DRQN, DPIQN, and DRPIQN.  The average rewards are evaluated over 100,000 episodes.   The types of the opponent agent's modes include ``hybrid'', ``offensive'', and ``defensive''.  A hybrid mode means that the opponent's mode is either offensive or defensive, and is determined randomly at the beginning of an episode.  Once the mode is determined, it remains fixed till the end of that episode. The second and third columns of Table~\ref{table::soccer_1vs1_reward} correspond to the opponent's modes in the training and testing phases, respectively.  All models are trained for 2 million timesteps, corresponding to 800 epochs in the training phase.  The highest average rewards in each column are marked in bold.
The results show that DPIQN and DRPIQN outperform DQN and DRQN in all cases under the same hyperparameter setting.  No matter which mode the opponent belongs to, DPIQN and DRPIQN agents are both able to score a goal for around 99\% of the episodes.  The results indicate that incorporating the policy features of the opponent into the Q-value learning module does help DPIQN and DRPIQN to derive better Q values, compared to those of DQN and DRQN. 
Another interesting observation from Table~\ref{table::soccer_1vs1_reward} is that DQN agents are also able to achieve sufficiently high average rewards, as long as the mode of the opponent remains the same in the training and testing phases. When DQN agents face unfamiliar opponents in the testing phase, they play poorly and lose the games most of the time.  This indicates that DQN agents are unable to adapt themselves to opponents with different strategies and, hence, non-stationary environments.  We have also observed that DPIQN and DRPIQN agents tend to play aggressively in most of the games, while DQN and DRQN agents are often confused by the opponent's moves. \\
\indent Fig.~\ref{figure::learning_curves_1vs1} plots the learning curves of the four models in the training phase. The numbers in Fig.~\ref{figure::learning_curves_1vs1} are averaged from the scores of the first 500 episodes in each epoch.  The opponent's modes in all of the four cases are random.  It can be seen that DPIQN and DRPIQN learn much faster than DQN and DRQN.  DRPIQN's curve increases slower than DPIQN's due to the extra parameters from the LSTM layers in DRPIQN's model.  Please note that the final average rewards of DPIQN and DRPIQN converge to around 0.8.  This is largely due to the $\epsilon$-greedy technique we used in the training phase.  Another reason is that in the testing phase, the average rewards are evaluated over 100,000 episodes, leading to less variations. 
%\begin{figure*}[tb]
%\begin{floatrow}[3]
%  \ffigbox[.9\FBwidth]{
%    \includegraphics[width=\linewidth]{mean_reward_1vs1.png}
%  }{
%   	\vspace{-1.3em}
%    \caption{Learning curve comparison in the 1 vs. 1 scenario.}
%    \label{figure::learning_curves_1vs1}
%  }
%  \ffigbox[.9\FBwidth]{
%    \includegraphics[width=\linewidth]	{mean_reward_2vs2_baseline.png}
%	}{
%  	\vspace{-2.4em}
%    \caption{Learning curve comparison in the 2 vs. 2 scenario.}
%    \label{figure::learning_curves_2vs2}
%  }
%  \ffigbox[.9\FBwidth]{
%  	\includegraphics[width=\linewidth]{mean_reward_1vs1_reg_lamb.png}
%  } {
%    \vspace{-2.4em}
%    \caption{Impact of $\lambda$ and $L^{PI}$ on average reward}
%    \label{figure::mean_reward_1vs1_reg_lamb}
%  }
  % Learning curves in 1vs1
  % Testing result in 1vs1
  %\caption{``Hybrid'' means opponent selects mode randomly in beginning of each episode; "Offensive" and "Defensive" mean opponent is fixed as offensive and defensive mode respectively. ``Training'' column specifies opponent setting during training. In ``Testing'' column, we report mean reward over 100,000 episodes with three different settings. The model with highest mean reward in each setting is marked as \textbf{bold}.}
%\end{floatrow}
%\label{figure::1vs1}
%\vspace{-1.3em}
%\end{figure*}

\subsection{Collaboration with Rule-based Agent}
\noindent Table~\ref{table::soccer_2vs2_reward} compares the average rewards of the controllable agent's team in the 2 vs. 2 scenario for four types of models used to implement the controllable agent.  Similarly, the average rewards are evaluated over 100,000 episodes.  Both the collaborator and opponents are rule-based agents, and are set to the hybrid mode.  The second column of Table~\ref{table::soccer_2vs2_reward} indicates which rule-based agent's policy features are learned by DPIQN and DRPIQN agents in the training phase.  ``Both'' means that DPIQN and DRPQIN agents learn the policy features of both the collaborator and the opponents, while ``C-only''/``O-only'' represents that only the policy features of the collaborator/the opponents are considered by our agents, respectively.  We denote them as \emph{DPIQN/DRPIQN (B)}, \emph{DPIQN/DRPIQN (C)}, and \emph{DPIQN/DRPIQN (O)} for these three different settings.  All models are trained for 3 million timesteps (1,200 epochs).  The highest average reward in the third column is marked in bold. \\
\indent From the results in Table~\ref{table::soccer_2vs2_reward}, we can see that DPIQN and DRPIQN agents are much superior to DQN and DRQN agents in the 2 vs 2 scenario.  For most of the episodes, DQN and DRQN agents' teams lose the game with their average rewards less than zero.  On the other hand, DPIQN and DRPIQN agents' teams are able to demonstrate a higher goal scoring ability.  We have observed in our experiments that our agents have learned to pass the ball to its teammate, or save the ball from its teammate chased by the opponents.  This implies that our agents had learned to collaborate with its teammate.  On the contrary, the baseline DQN and DRQN agents are often confused by the opponent team's moves, and are relatively conservative in deciding their actions.  As a result, they usually stand still in the same place, resulting in losing the possession of the ball. Lastly, it is noteworthy that both DRPIQN (C) and DPIQN (O) achieve high average rewards. This indicates that for our models, it is sufficient to only model a subset of agents in the environment. Therefore, we consider DPIQN and DRPIQN to be potentially scalable to a more complex environment.\\
\indent Fig.~\ref{figure::learning_curves_2vs2} plots the learning curves of DPIQN (B), DRPIQN (B), DQN, and DRQN in the training phase.  Similarly, the numbers in Fig.~\ref{figure::learning_curves_2vs2} are averaged from the scores of the first 500 episodes in each epoch.  It can again be observed that the learning curves of DPIQN and DRPIQN grow much faster than those of DQN and DRQN.   Even at the end of the training phase, the average rewards of our models are still increasing.  From the results in Table~\ref{table::soccer_2vs2_reward} and Fig.~\ref{figure::learning_curves_2vs2}, we conclude that DPIQN and DRPIQN are generalizable to complex environments with multiple agents. 

\subsection{Collaboration with Learning Agent}
\begin{figure}[!tb]
	\includegraphics[width=0.85\linewidth]{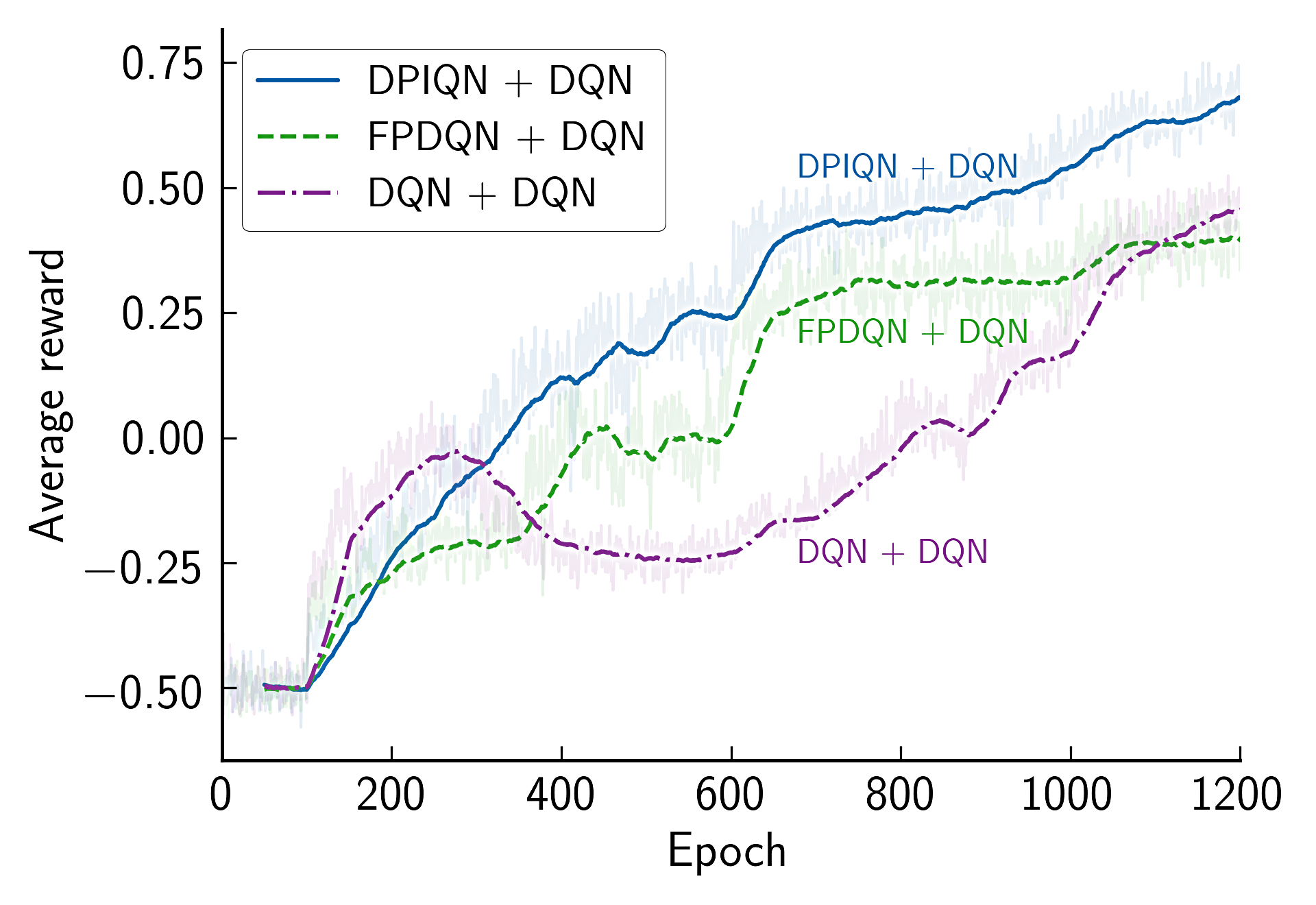}
    \vspace{-0.5em}
	\captionof{figure}{Mean rewards of different collaboration teams.}
	\label{figure::mean_reward_2vs2_collab}
    \vspace{-1.5em}
\end{figure}
\noindent To test our model's ability to cooperate with a learning collaborator, we conduct further experiments in the 2 vs. 2 scenario. As the environment contains two learning agents in this setting, we additionally introduce fingerprints DQN (FPDQN)~\footnote{FPDQN takes as input {$(s, \epsilon, e)$}, where {$e$} is the current training epoch, as suggested in the original paper.}~\cite{foerster2017stabilising}, a DQN-based multi-agent RL method to tackle non-stationarity, as a baseline model. Each of the three types of models, including DQN, FPDQN and DPIQN, has to team up with a learning DQN agent to play against the opposing team. All models mentioned above adopt the same parameter settings, and the rule-based agents in the opposing team are set to the hybrid mode. The average rewards are similarly evaluated over 100,000 episodes. Please note that DPIQN only models its collaborator, i.e. the DQN agent, in the experiment.

\begin{table}[!tb]
	\caption{Evaluation result of 2 vs. 2 scenario with a learning \\collaborator.}
	\label{table::soccer_2vs2_learning_agent}
	\centering
	\begin{tabular}{l|c|c|C{6.5em}}
	\Xhline{2\arrayrulewidth}
		Model   & Scoring Rate & Draws    & Avg. rewards       \\ \hline
    	DPIQN 	&  31.54\% & \textbf{11.97\%} & \textbf{0.829} \\
    	FPDQN 	&  0.00\% & 36.83\% & 0.555\\
    	DQN 	&  58.70\% & 39.77\% & 0.529\\
	\Xhline{2\arrayrulewidth}
	\end{tabular}
    %\vspace{-1.5em}
\end{table}

\begin{table}[!tb]
	\caption{Impact of unfamiliar agents on average reward.}
	\label{table::unseen_reward}
	\centering
	\small
	\begin{tabular}{p{3.5em}cC{5.5em}C{5.5em}}
	\Xhline{2\arrayrulewidth}
	\multicolumn{2}{c}{}        & Unfamiliar-O                & Unfamiliar-C                                         \\ \hline \hline
	\multicolumn{4}{c}{1 vs. 1 Scenario}                                                                             \\ \hline
	\multicolumn{2}{l|}{DPIQN}                                & 0.909 (90\%)                  & -                    \\
	\multicolumn{2}{l|}{DRPIQN}                               & \textbf{0.947 (94\%)}         & -                    \\ \hline \hline
	\multicolumn{4}{c}{2 vs. 2 Scenario (rule-based)}                                                                              \\ \hline
	\multirow{3}{*}{DPIQN}      & \multicolumn{1}{c|}{Both}   & 0.501 (65\%)                  & 0.645 (84\%)          \\
                            & \multicolumn{1}{c|}{O-only} & 0.488 (75\%)                  & 0.535 (82\%)         \\
                            & \multicolumn{1}{c|}{C-only} & 0.076 (32\%)                  & 0.189 (81\%)         \\
	\multirow{3}{*}{DRPIQN}     & \multicolumn{1}{c|}{Both}   & 0.534 (76\%)                  & 0.565 (81\%)         \\
                            & \multicolumn{1}{c|}{O-only} & \textbf{0.578 (80\%)}         & \textbf{0.625 (87\%)} \\
                            & \multicolumn{1}{c|}{C-only} & 0.625 (73\%)                  & 0.695 (82\%)          \\
	\Xhline{2\arrayrulewidth}
	\end{tabular}
    %\vspace{-0.5em}
\end{table}

%\begin{table*}[tb]
%	\centering
	% Mean reward: 2vs2 collaboration
	% Mean reward: 1vs1 reg lamb
%	\begin{minipage}[t]{0.3\textwidth}
%		\includegraphics[width=\linewidth]{mean_reward_1vs1_reg_lamb.png}
%			\captionof{figure}{Impact of $\lambda$ and $L^{PI}$ on average reward.}
%			\label{figure::mean_reward_1vs1_reg_lamb}
%	\end{minipage}
%	\hfill
	% Q cost: 1vs1 reg lamb
%	\begin{minipage}[t]{0.3\textwidth}
%		\includegraphics[width=\linewidth]{q_cost_1vs1_reg_lamb.png}
%			\captionof{figure}{Impact of $\lambda$ and $L^{PI}$ on $L^Q$.}
%			\label{figure::loss_2vs2_reg_q_cost}
%	\end{minipage}
 %   \vspace{-1em}
%\end{table*}
Fig.~\ref{figure::mean_reward_2vs2_collab} plots the learning curves of the three teams in the training phase. The numbers in Fig.~\ref{figure::mean_reward_2vs2_collab} are similarly averaged from the scores of the first 500 episodes in each epoch. It can be observed that the learning curves of both DPIQN and FPDQN grow much faster and steadier than DQN. It takes DQN 1,000 epochs to reach the same level of performance as FPDQN. On the other hand, DPIQN consistently receives higher average rewards than the other models. From the results in Fig.~\ref{figure::mean_reward_2vs2_collab}, we show the effectiveness of our model in a multi-agent learning setting.

In Table~\ref{table::soccer_2vs2_learning_agent}, we report the scoring rate for each controllable agent, the percentage of draws, as well as the evaluation results in the testing phase. We observe that the DPIQN agent only scores 31.54\% of the time, and runs forward to support its teammate's attack whenever the collaborator gains possession of the ball and advances towards the goal. On the contrary, when DPIQN agent possesses the ball, we do not observe any similar behavior from the DQN collaborator. The result shows that DPIQN is superior in cooperating with and assisting its teammate through modeling the collaborator. Furthermore, although FPDQN learns faster than DQN agents' team, it does not seem to show any sign of collaboration with its teammate, and spends most of its time staying in the offensive half. Therefore, it achieves only a 0.00\% scoring rate despite its higher average rewards than the DQN team. We also observe that the DQN agents' team hardly ever steals the ball from the rule-based agents, and therefore often ends up in draws. In addition, when one of the DQN agents possesses the ball, the other DQN tends to stand still, instead of assisting its teammate. This behavior is similar to what we have observed in DPIQN's team. We therefore conclude that DQN agents lack the ability to cooperate efficiently. From the result in Figure~\ref{figure::mean_reward_2vs2_collab} and Table~\ref{table::soccer_2vs2_learning_agent}, we further show that our proposed model is capable of handling non-stationarity, as well as improving the overall performance of RL agents.

\begin{figure}[!tb]
	\includegraphics[width=0.85\linewidth]{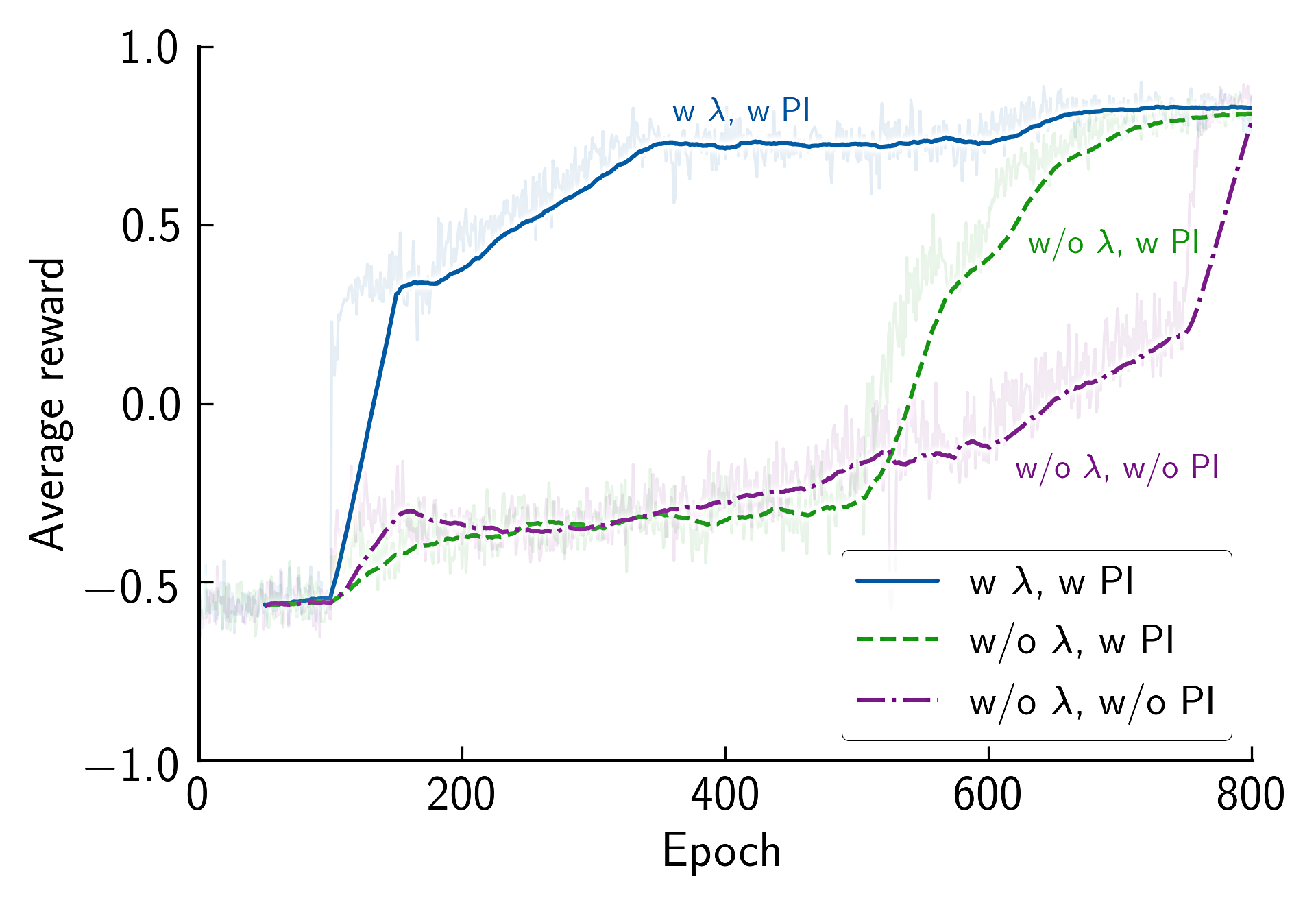}
    \vspace{-0.5em}
	\captionof{figure}{Impact of $\lambda$ and $L^{PI}$ on average reward.}
	\label{figure::mean_reward_1vs1_reg_lamb}
    \vspace{-1.5em}
\end{figure}
\subsection{Generalizability to Unfamiliar Agent}
\noindent In this section, we show that DPIQN and DRPIQN are capable of dealing with unfamiliar agents whose policies change over time.  We evaluate our models in the two scenarios discussed above, and summarize our results in Table~\ref{table::unseen_reward}.  In the training phase, DPIQN and DRPIQN agents are trained against rule-based agents with fixed policies in an episode.  However, in the testing phase, the policies of the collaborator or opponents are no longer fixed.  Each of them may randomly updates its policy mode (either offensive of defensive) with an irregular update period ranging from 4 to 10 timesteps.  The update period is also randomly determined.  This setting thus makes the controllable agent unfamiliar with the policies of its collaborator or opponents in the testing phase, allowing us to validate the generalizability of DPIQN and DRPIQN.  Note that the average rewards listed in Table~\ref{table::unseen_reward} are evaluated over 100,000 episodes. \\\indent Table ~\ref{table::unseen_reward} compares two cases for validating the generalizability of DPIQN and DRPIQN:  unfamiliar opponents (abbreviated as unfamiliar-O) and unfamiliar collaborator (abbreviated as unfamiliar-C).  These two cases correspond to the second and third columns of Table ~\ref{table::unseen_reward}, respectively.  We report results for the 1 vs. 1 and 2 vs. 2 scenarios in separate rows.  The numbers in the parenthesis are ratios of the average rewards in Table~\ref{table::unseen_reward} to those of the corresponding entries in Tables~\ref{table::soccer_1vs1_reward}~\&~\ref{table::soccer_2vs2_reward}.  In the 1 vs. 1 scenario, DPIQN and DRPIQN are able to achieve average rewards of 0.909 and 0.947, respectively, even when confronted with an unfamiliar opponent.  In the 2 vs. 2 scenario, both DPIQN and DRPIQN maintain their performance in most of the cases.  It can be seen that DRPIQN performs slightly better than DPIQN in these two scenarios when playing against unfamiliar opponents.  We have observed that DRPIQN (O) achieves the highest average reward ratios among all of the cases.  One potential explanation is that DRPIQN (O) focuses only the policy features of the opponents in the training phase, therefore is able to adapt itself to unfamiliar opponents better than the other settings. Table~\ref{table::unseen_reward} also indicates that DPIQN and DRPIQN perform better when facing with unfamiliar collaborators than unfamiliar opponents. We have observed that when collaborating with an unfamiliar agent, DPIQN and DRPIQN agents tend to score a goal by itself, due to its lack of knowledge about the collaborator's intentions.
\subsection{Ablative Analysis}
\begin{figure}[!tb]
	\includegraphics[width=0.85\linewidth]{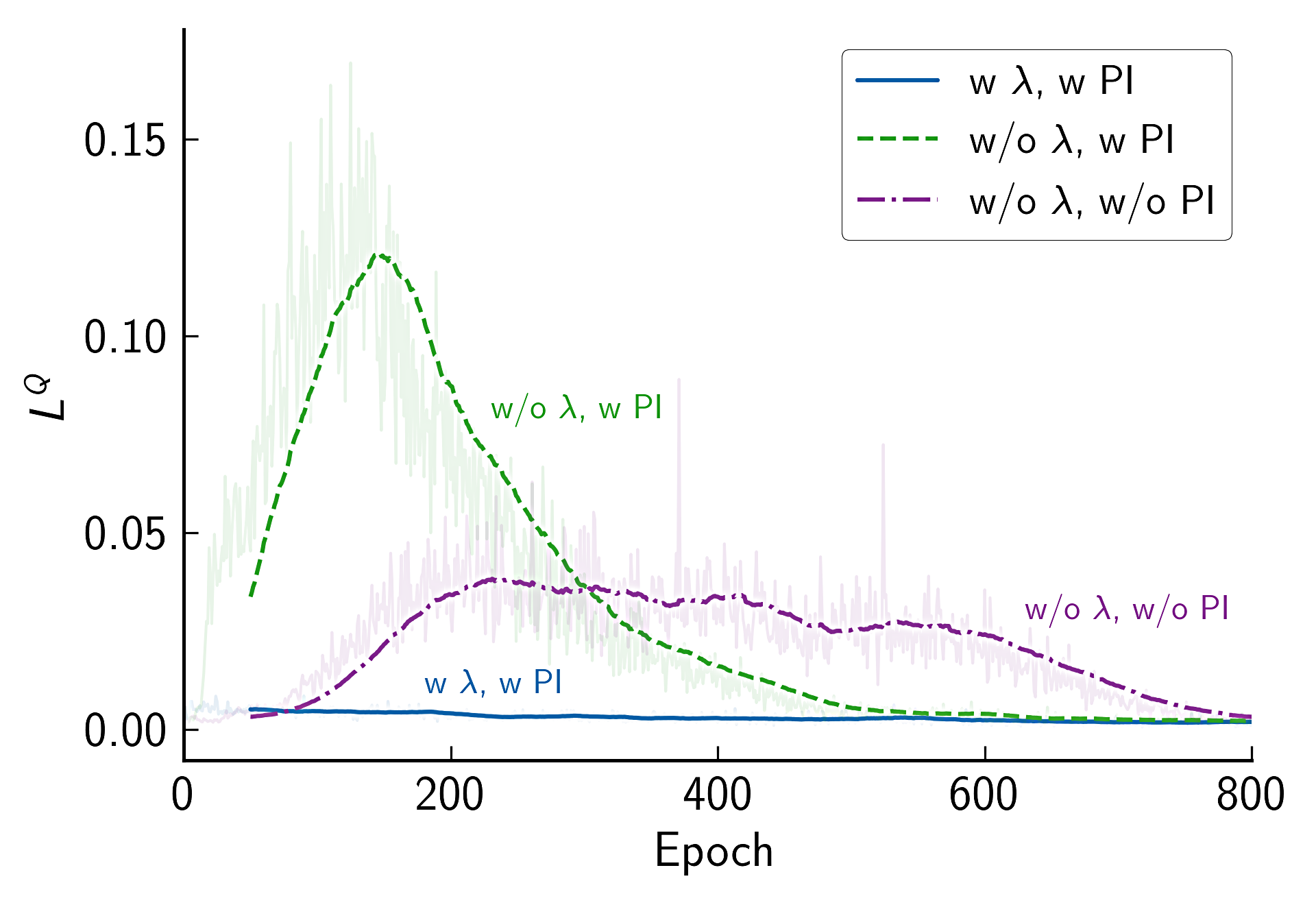}
    \vspace{-0.5em}
	\captionof{figure}{Impact of $\lambda$ and $L^{PI}$ on $L^Q$.}
	\label{figure::loss_2vs2_reg_q_cost}
   	\vspace{-1.5em}
\end{figure}

We further investigate the effectiveness of our adaptive loss function by a detailed analysis of $L^{PI}$ and $L^Q$.  Moreover, we plot the learning curves of three different cases, and show that our adaptive loss design helps accelerate convergence and stabilize training.  We focus exclusively on DPIQN, as DRPIQN produces similar results. \\
\indent Fig.~\ref{figure::mean_reward_1vs1_reg_lamb} illustrates the learning curves of DPIQN in the 1 vs. 1 scenario.  These three curves correspond to DPIQN models trained with or without the use of $\lambda$ and $L^{PI}$ in Eq.~\ref{eq::dpiqn_loss}.  Although all of the three cases converge to an average reward of 0.8 at the end, the one trained with both $\lambda$ and $L^{PI}$ converges much faster than the others.  We further analyze $L^Q$ for the three cases in Fig.~\ref{figure::loss_2vs2_reg_q_cost}.  It is observed that the DPIQN model trained with both $\lambda$ and $L^{PI}$ shows less fluctuations in $L^Q$ and thus better stability than the other two cases in the training phase.  We conclude that both policy inference and adaptive loss are essential to DPIQN's performance.
\section{Conclusion}\label{sec.conclusion}
\noindent  In this paper, we presented an in-depth design of DPIQN and its variant DRPIQN, suited to multi-agent environments.  We presented the concept of policy features, and proposed to incorporate them as a hidden vector into the Q-networks of the controllable agent.  We trained our models with an adaptive loss function, which guides our models to learn policy features before Q values.   We extended the architectures of DPIQN and DRPIQN to model multiple agents, such that it is able to capture the behaviors of the other agents in the environment. We performed experiments for two soccer game scenarios, and demonstrated that DPIQN and DRPIQN outperform DQN and DRQN in various settings. Moreover, we verified that DPIQN is capable of dealing with non-stationarity by conducting experiments where the controllable agent has to cooperate with a learning agent, and showed that DPIQN is superior in collaboration to a recent multi-agent RL approach. We further validated the generalizability of our models in handling unfamiliar collaborators and opponents.  Finally, we analyzed the loss function terms, and demonstrated that our adaptive loss function does improve the stability and learning speed of our models.

\section{Acknowledgments}
\label{Acknowledgments}
The authors thank MediaTek Inc. for their support in researching funding, and NVIDIA Corporation for the donation of the Titan X Pascal GPU used for this research.

%%%%%%%%%%%%%%%%%%%%%%%%%%%%%%%%%%%%%%%%%%%%%%%%%%%%%%%%%%%%%%%%%%%%%%%%%%%%%%%%%%%%%%%%%%%%%%%%%%%%%%%%%
%% bibliography: see CFP for number of permitted pages

\bibliographystyle{ACM-Reference-Format}  % do not change this line!
\bibliography{reference}  % put name of your .bib file here

\end{document}